%% file: samplepaper.tex
\documentclass[runningheads]{llncs}
\usepackage[T1]{fontenc}
\usepackage{graphicx}
\usepackage{amsmath}
\usepackage[table]{xcolor}
\usepackage{booktabs}
\usepackage{multirow}
\usepackage{tabularx}
\usepackage{caption}
\usepackage{subcaption}
\usepackage{makecell}

\makeatletter
\newcommand{\printfnsymbol}[1]{%
  \textsuperscript{\@fnsymbol{#1}}%
}
\makeatother

\begin{document}
\title{A Surface-normal Based Neural Framework for Colonoscopy Reconstruction}
%
%

%
%
%

\author{Shuxian Wang\thanks{These authors contributed equally to this work}\textsuperscript{1} \and
Yubo Zhang\printfnsymbol{1}\textsuperscript{1} \and
Sarah K. McGill\textsuperscript{2} \and
Julian G. Rosenman\textsuperscript{3} \and
Jan-Michael Frahm\textsuperscript{1} \and
Soumyadip Sengupta\textsuperscript{1} \and
Stephen M. Pizer\textsuperscript{1}
}
\authorrunning{S. Wang and Y. Zhang et al.}
%
\institute{University of North Carolina at Chapel Hill \\
\textsuperscript{1}\email{\{shuxian,zhangyb,jmf,ronisen,pizer\}@cs.unc.edu} \\
\textsuperscript{2}\email{mcgills@email.unc.edu}  \quad 
\textsuperscript{3}\email{rosenmju@med.unc.edu}
}
\maketitle              
\begin{abstract}
Reconstructing a 3D surface from colonoscopy video is challenging due to illumination and reflectivity variation in the video frame that can cause defective shape predictions.
Aiming to overcome this challenge, we utilize the characteristics of surface normal vectors and develop a two-step neural framework that significantly improves the colonoscopy reconstruction quality.
The normal-based depth initialization network trained with self-supervised normal consistency loss provides depth map initialization to the normal-depth refinement module, which utilizes the relationship between illumination and surface normals to refine the frame-wise normal and depth predictions recursively.
Our framework's depth accuracy performance on phantom colonoscopy data demonstrates the value of exploiting the surface normals in colonoscopy reconstruction, especially on en face views.
Due to its low depth error, the prediction result from our framework will require limited post-processing to be clinically applicable for real-time colonoscopy reconstruction.

\end{abstract}

\input{0_intro}
\input{1_background.tex}
\input{2.5_methods.tex}
\input{4_experiments}
\input{5_conclusion}


%
%
%
\bibliographystyle{splncs04}
\bibliography{ref}

\end{document}

%% file: 0_intro.tex
\section{Introduction}
\begin{figure}[t]
    \centering
    \includegraphics[width=0.9\textwidth]{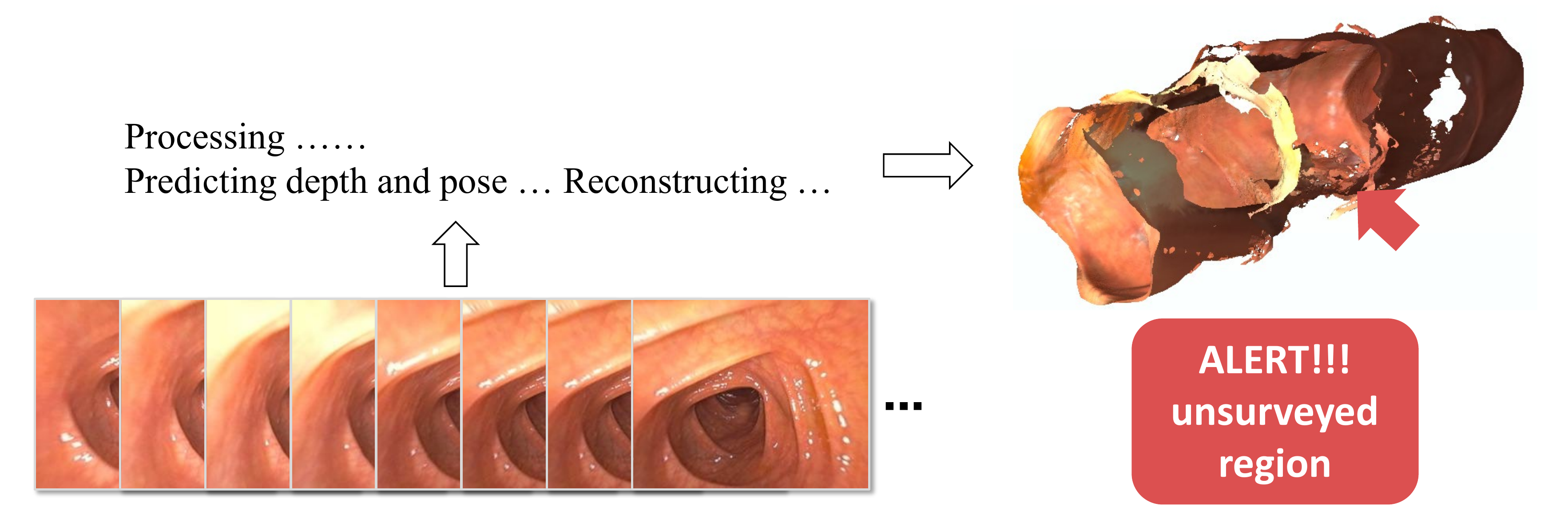}
    \vspace{-1em}
    \caption{Reconstructing the 3D mesh from a colonoscopy video in real-time according to the predicted depth and camera pose, allowing holes in the mesh to alert the physician to unsurveyed regions on the colon surface.}
    \label{fig:intro}
    \vspace{-1.5em}
\end{figure}

Reconstructing the 3D model of colon surfaces concurrently during colonoscopy improves polyp (lesion) detection rate by lowering the percentage of the colon surface that is missed during examination~\cite{hong2007colonoscopy}.
Often surface regions are missed due to oblique camera orientations or occlusion by the colon folds.
By reconstructing the surveyed region, the unsurveyed part can be reported to the physician as holes in the 3D surface (as in Fig.~\ref{fig:intro}).
This approach makes it possible to guide the physician back and examine the missing region without delay.

To reconstruct colon surfaces from the colonoscopy video, a dense depth map and camera position need to be predicted from each frame. 
Previous work \cite{liu2019dense,ma2019real} trained deep neural networks to predict the needed information in real time. 
With the proper help from post-processing~\cite{liu2020reconstructing,ma2021rnnslam}, these methods often are able to reconstruct frames with abundant photometric and geometric features such as in  "down-the-barrel" (axial) views where the optical axis is aligned with the organ axis. However, they often fail to reconstruct from frames where the optical axis is perpendicular to the surface ("en face" views). We address the problem of reconstruction from these en face views.
In our target colonoscopy application, the geometry of scenes in these two viewpoints are significantly different, manifesting as a difference in depth ranges.
In particular, the en face views have near planar geometry, resulting in limited geometric structures informing the photometric cues. As a result, dense depth estimation is challenging using photometric cues alone. However, the characteristics of the endoscopic environment (with a co-located light source and camera located in close proximity to the highly reflective mucus layer coating the colon) mean that illumination is a strong cue for understanding depth and the surface geometry. We capitalize upon this signal to improve reconstruction in en face views.
We also aim to yield the reconstruction from frame-wise predictions with minimal post-integration to achieve near real-time execution, which requires strong geometric awareness of the network.

In this work we build a neural framework that fully exploits the surface normal information for colonoscopy reconstruction.
Our approach is two-fold, 1) {\bf normal-based depth initialization} (Section \ref{sec:init}) followed by 2) {\bf normal-depth refinement} (Section \ref{sec:nr}).
Trained with a large amount of clinical data, the normal-based depth initialization network alone can already provide good-quality reconstructions of ``down-the-barrel'' video segments.
To improve the performance on en face views, we introduced the normal-depth refinement module to refine the depth prediction.
We find that the incorporation of surface normal-aware losses improves both frame-wise depth estimation and 3D surface reconstruction from the C3DV~\cite{bobrow2022} and clinical datasets, as indicated by both measurements and visualization.

%% file: 1_background.tex
\section{Background} \label{sec:background}
Here we describe prior work on 3D reconstruction from endoscopic video, particularly focusing on colonoscopic applications. 
They usually start with a neural module to provide frame-wise depth and camera pose estimation, followed by an integration step that combines features across a video sequence to generate a 3D surface.
With no ground truth from clinical data to supervise the frame-wise estimation network training, some methods transferred the prior learned from synthetic data to real data~\cite{cheng2021depth,mahmood2018unsupervised,mathew2020augmenting} while
others utilized the self-consistent nature of video frames to conduct unsupervised training~\cite{liu2019dense}.
In order to incorporate optimization-based methods to calibrate the results from learning-based methods,
Ma et al.~\cite{ma2019real,ma2021rnnslam} introduced the system with a SLAM component~\cite{engel2017direct} and a post-averaging step to correct potential camera pose errors;
Bae et al.~\cite{bae2020deep} and Liu et al.~\cite{liu2020reconstructing} integrated Structure-from-Motion~\cite{schonberger2016structure} with the network, trading off time efficiency for better dense depth quality.

When using widely-applied photometric and simple depth consistency objectives~\cite{bian2019unsupervised,zhou2017unsupervised} in training, networks frequently fail to predict high quality and temporally consistent results due to the low geometric texture of endoscopic surfaces and time-varying lighting~\cite{zhang2021lighting}. The corresponding reconstructions produced by these methods have misaligned or unrealistic shapes as a result.
Meanwhile, recent work in computer vision has shown surface normals to be useful for enforcing additional geometric constraints in refining depth predictions~\cite{li2021structdepth,yang2018lego,yu2022monosdf} while the relationship between surface normals and scene illumination has been exploited in photometric stereo~\cite{li2018learningrecons,lichy2022fast,lichy2021shape,xie2019mesh}.
The success of utilizing surface normals in complex scene reconstruction inspires us to explore this property in the endoscopic environment.

%% file: 2.5_methods.tex
\section{Methods} \label{sec:methods}
\begin{figure}[t]
    \centering
    \includegraphics[width=0.99\textwidth]{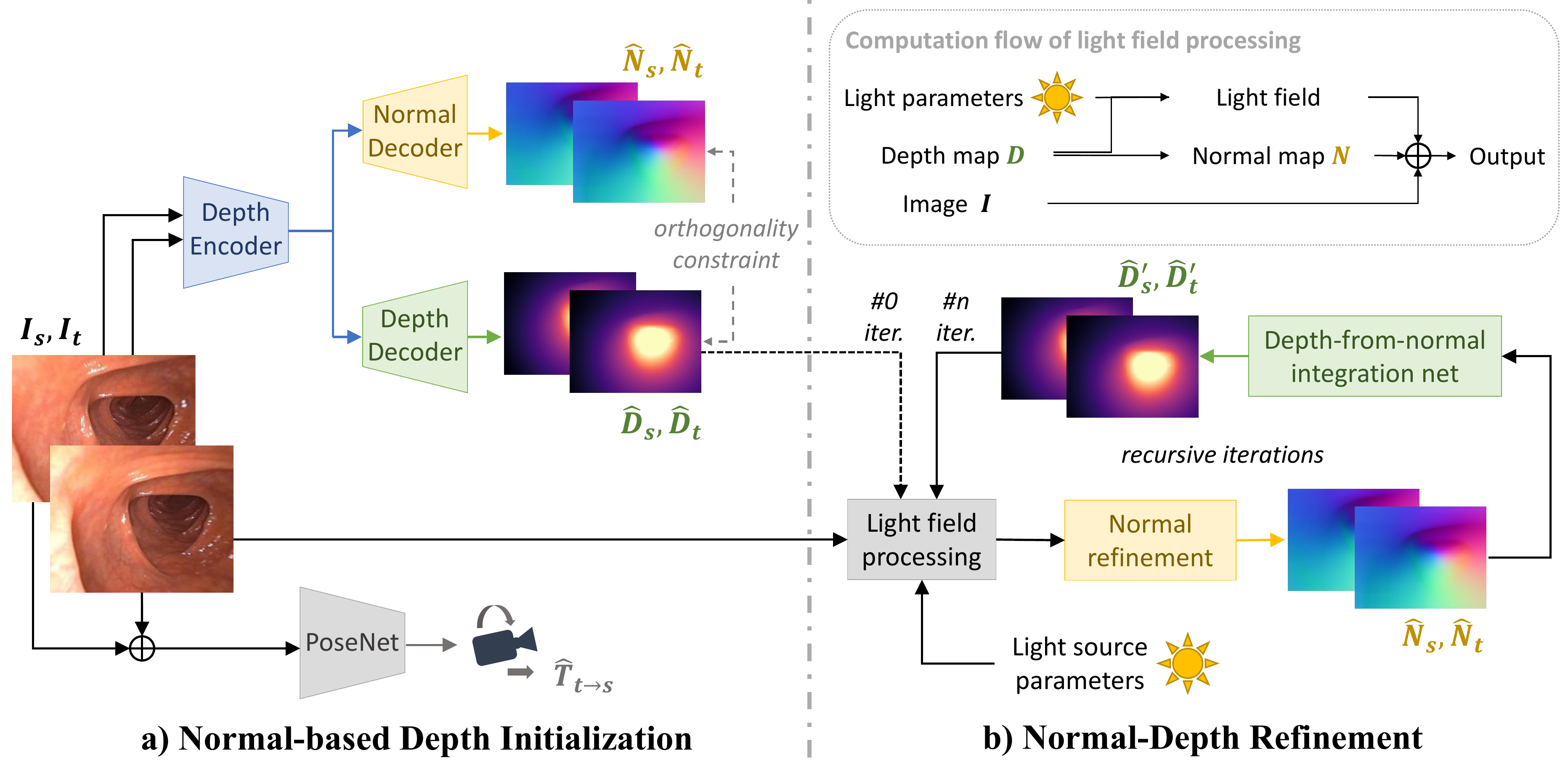}
    \vspace{-0.5em}
    \caption{Our two-fold framework of colonoscopy reconstruction. a) {\bf Normal-based depth initialization} network is trained with self-supervised surface normal consistency loss to produce depth map and camera pose initialization. b) {\bf Normal-depth refinement} framework utilizes the relation between illumination and surface geometry to refine depth and normal predictions.}
    \label{fig:network}
    \vspace{-1.5em}
\end{figure}

Surface normal maps describe the orientation of the 3D surface and reflect local shape knowledge.
We incorporate this information in two ways: first, to enhance unsupervised consistency losses in our normal-based depth initialization (Fig. \ref{fig:network}a) and second, to allow us to use illumination information in our normal-depth refinement (Fig. \ref{fig:network}b). We use this framework as initialization for a SLAM-based pipeline that fuses the frame-wise output into a 3D mesh following Ma et al.~\cite{ma2021rnnslam}.

\subsection{Normal-based Depth Initialization} \label{sec:init}
In order to fully utilize the large amount of unlabeled clinical data, our initialization network is trained with self-supervision signals based on the scene's consistency of frames from the same video.
We particularly exploit the surface normal consistency in training to deal with the challenges of complicated colon topology in addition to applying the commonly used photometric consistency losses~\cite{bian2019unsupervised,monodepth2,zhou2017unsupervised}, which are less reliable due to lighting complexity in our application.
Trained with the scheme described below, this network produces good depth and camera pose initialization for later reconstruction. We refer to this model as "NormDepth" or "ND" in Section \ref{sec:experiments}.

\paragraph{Background - projection}
The self-supervised training losses discussed in this section are built upon the pinhole camera model and the projection relation between a source view $s$ and a target view $t$~\cite{zhou2017unsupervised}.
Given the camera intrinsic $K$, a pixel $p_t$ in a target view can be projected into the source view according to the predicted depth map $\hat D_t$ and the relative camera transformation $\hat T_{t \to s}$.
This process yields the pixel's homogeneous coordinates $\hat p_s$ and its projected depth $\hat d_s^t$ in the source view, as in Eq.~\ref{Eq.proj}:
\begin{gather}
    \label{Eq.proj}
    \hat p_s, \hat d^t_s \sim K \hat T_{t \to s} \hat D_t(p_t) K^{-1} p_t
\end{gather}

\subsubsection{Normal Consistency Objective}
\label{sec:norm_consist}
As the derivative of vertices' 3D positions, surface normals can be sensitive to the error and noise on the predicted surface.
Therefore, when the surface normal information is appropriately connected with the network's primary predictions, i.e., the depth and camera pose, utilizing surface normal consistency during training can further correct the predictions and improve the shape consistency.

Let $\hat N_t$ be the object's surface normals in the target coordinate system.
In the source view's coordinate system, the direction of those vectors depends on the relative camera rotation $\hat R_{t \to s}$ (the rotation component of $\hat T_{t \to s}$) and should agree with the source view's own normal prediction $\hat N_s$; using this correspondence we form the normal consistency objective as
\begin{gather}
\label{Eq.normal}
    L_{norm} = || \hat N_s \left \langle \hat p_s \right \rangle - \hat R_{t \to s} \hat N_t ||_1
\end{gather}
Here, we use the numerical difference between the two vectors (L1 loss) for error.
In practice, we find that using angular difference has similar performance.

\paragraph{Surface Normal Prediction}
We found that when training with colonoscopy data, computing normals directly from depths as in some previous work~\cite{yang2018lego,yang2018unsupervised} is less stable and tends to result in unrealistic shapes.
Instead, we built the network to output the initial surface normal information individually, and trained it in consensus with depth prediction using $L_{orth}$:
\begin{gather}
\label{Eq.orth}
    \hat V(p) = \hat D(p_a) K^{-1} p_a - \hat D(p_b) K^{-1} p_b \\
    L_{orth} = \sum_p \hat N(p) \cdot \hat V(p)
\end{gather}
where $\hat V(p)$ is the approximate surface vector around $p$, which is computed from the depths of $p_a$ and $p_b$, $p$'s nearby pixels.
In practice, we apply two pairs of $p_{a/b}$ position combinations, i.e., $p$'s top-left/bottom-right and top-right/bottom-left neighboring pixels.
This orthogonality constraint bridges the surface normal and depth outputs so that the geometric consistency constraint on the normal will in turn regularize the depth prediction.

\subsubsection{Training Overview}
We adapt our depth initialization network from Godard et al.~\cite{monodepth2} with an additional decoder to produce per-pixel normal vectors besides depths, and apply their implementation of photometric consistency loss $L_{photo}$ and depth smoothness loss $L_{sm}$.
Besides the surface normal consistency, we also enforce the prediction's geometric consistency by minimizing the difference between the predicted depths of the same scene in different frames, as in~\cite{bian2019unsupervised}:
\begin{gather}
\label{Eq.depth}
    L_{depth} = \frac{\left| \hat D_s \left \langle \hat p_s \right \rangle - \hat D^t_s \right|}{\hat D_s \left \langle \hat p_s \right \rangle + \hat D^t_s}
\end{gather}
With the per-pixel mask $M$ to mask out the stationary~\cite{monodepth2}, invalid projected or specular pixels, the final training loss to supervise this initialization network is the weighted sum of the above elements, where $\lambda_{1-4}$ are the hyper-parameters:
\begin{equation}
\begin{aligned}
    L^{init} =& (L_{photo} + \lambda_1 L_{norm} + \lambda_2 L_{depth}) \odot M 
    + \lambda_3 L_{orth} + \lambda_4 L_{sm}
\end{aligned}
\label{Eq.init-loss}
\end{equation}

\subsection{Normal-Depth Refinement} \label{sec:nr}

In the endoscopic environment, there is a strong correlation between the scene illumination from the point light source and the scene geometry characterized by the surface normals.
Our normal-depth refinement framework uses a combination of the color image, scene illumination as represented by the light field, and an initial surface normal map as input. We use both supervised and self-supervised consistency losses to simultaneously enforce improved normal map refinement and consistent performance across varying scene illumination scenarios.

\paragraph{Light Field Computation}
We use the light field to approximate the amount of light each point on the viewed surface receives from the light source. As in Lichy et al. ~\cite{lichy2022fast} we parameterize our light source by its position relative to the camera, light direction, and angular attenuation. In the endoscopic environment, the light source and camera are effectively co-located so we take the light source position and light direction to be fixed at the origin $O$ and parallel to the optical axis $\vec{z}$, respectively. Thus for attenuation $\mu$ and depth map $\hat{D}$, we define the point-wise light field $\hat{F}$ and the point-wise attenuation $\hat{A}$ as
\begin{equation}
    \hat{F} = \frac{O - \hat{D}}{|| O - \hat{D} ||}, \quad
    \hat{A} = \frac{(-\sum \hat{F} \cdot \vec{z})^\mu}{|| O - \hat{D} ||^2}
\end{equation}
For our model input, we concatenate the RGB image, $\hat{F}$, $\hat{A}$, and normal map $\hat{N}$ (computed from the gradient of the depth map) along the channel dimension.

\subsubsection{Training Overview}
\label{sec:nr_training}

In order to use illumination in colonoscopy reconstruction, we  adapt our depth-normal refinement model from Lichy et al.~\cite{lichy2022fast} with additional consistency losses and modified initialization.
We use repeated iterations for refinement; in order to reduce introduced noise, we use a multi-scale network as in many works in neural photometric stereo~\cite{li2018learningrecons,lichy2022fast,lichy2021shape}. After each recursive iteration, we upsample the depth map to compute normal refinement at a higher resolution. We denote $n$ iterations with "$n \times$NR".

We compute the following losses for each scale, rescaling the ground truth where necessary to match the model output. For the supervised loss $L_{gt}$ for iteration $i$, we minimize L1 loss on the normal refinement module output $\hat{N}_i$ and the matching ground truth normal map $N$ as well as the L1 loss on the depth-from-normal model output $\hat{D}_i$ and the matching ground truth depth map $D$. We define a scaling factor $\alpha_i = \frac{\text{median}(D)}{\text{median}(\hat{D}_i)}$.
\begin{equation}
    L_{gt} = \sum_i || N - \hat{N}_i ||_1 + || D - \alpha_i \hat{D}_i ||_1
\end{equation}
For the depth-from-normal integration module, we compute a normal map $\hat{N}'_i$ from its depth output and minimize L1 loss between it and the input normal map $\hat{N}_i$; this has the effect of imposing the orthogonality constraint between the depth and surface normal maps.
\begin{equation}
    L_{dfn} = \sum_i || \hat{N}'_i - \hat{N}_i ||_1
\end{equation}

We use a multi-phase training regime for stability. 
In an iteration, we first train the normal refinement module and substitute an analytical depth-from-normal integration method. For the second phase, we freeze the normal refinement module and train only the depth-from-normal integration module.
For the third and final phase of training, we use the normal refinement module and neural integration, optimizing a weighted sum of all losses with hyperparameters $\lambda_1$ and $\lambda_2$. Thus we define the losses for each phase respectively as follows:
\begin{align}
    L_{refine}^{(1)} &= L_{gt} + \lambda_1 L_{norm} \\
    L_{refine}^{(2)} &= L_{dfn} \\
    L_{refine}^{(3)} &= L_{gt} + \lambda_1 L_{norm} + \lambda_2 L_{dfn}
\end{align}

%% file: 4_experiments.tex
\section{Experiments} \label{sec:experiments}

In our experiments, we demonstrate that incorporating surface normal information improves both frame-wise depth estimation and 3D surface reconstruction. 
We describe the frame-wise depth map improvement over baseline and the effect of various ablations in Section \ref{sec:ex-quant}.
To evaluate the effect of the frame-wise depth estimation on surface reconstruction, we compare the reconstructions obtained from initializing the SLAM pipeline~\cite{ma2021rnnslam} with the outputs from various methods of frame-wise depth estimation against initialization with ground truth depth maps. We provide a comparison of Chamfer distance \cite{open3d} on aligned mesh reconstructions in Table \ref{tab:cross_val} and a qualitative comparison in Section \ref{sec:ex-qual}. We also provide a qualitative comparison of the surfaces reconstructed from clinical video in Section \ref{sec:ex-clinical}.

\begin{table}[!b]
    \small
    \centering
    \vspace{-10pt}
    \begin{tabularx}{\textwidth}{c| >{\centering}p{0.13\textwidth} >{\centering}p{0.13\textwidth} >{\centering}p{0.13\textwidth} >{\centering}p{0.13\textwidth} |c}
        \multirow{2}{*}{\textbf{Method}} & \multicolumn{4}{c|}{\textbf{Depth Error} $\downarrow$} & \multirow{2}{*}{\textbf{{\makecell{Chamfer \\ Distance}}} $\downarrow$} \\
        \cline{2-5}
         & Abs Rel & Sq Rel & RMSE & log RMSE & \\
         \hline
        Monodepth2 \cite{monodepth2} & 0.189 & 2.878 & 11.779 & 0.232 & 0.057 $\pm$ 0.039 \\
        \hline
        NormDepth $- L_{norm}$ & 0.137 & 1.328 & 7.411 & \textbf{0.168} & 0.044 $\pm$ 0.018 \\
        NormDepth & 0.141 & 1.373 & 7.447 & 0.173 & 0.046 $\pm$ 0.019 \\
        \hline
        ND init $+\:1 \times$NR & \textbf{0.136} & \textbf{1.271} & \textbf{7.376} & 0.170 & \textbf{0.038} $\pm$ \textbf{0.017} \\
        ND init $+\:4 \times$NR & 0.141 & 1.353 & 7.479 & 0.173 & 0.044 $\pm$ 0.018 \\
        flat init $+\:4 \times$NR & 0.166 & 1.927 & 8.955 & 0.201 & 0.047 $\pm$ 0.022\\
        
    \end{tabularx}
    \vspace{3pt}
    \caption{Error averaged over 5-fold cross validation test sets of C3DV, $\pm$ standard deviation. Best performance in bold. "NormDepth" and "ND" stand for normal-based depth initialization and "$n\times$NR" stands for normal-depth refinement for $n$ iterations. "NormDepth $- L_{norm}$" denotes NormDepth trained without $L_{norm}$. "flat init" denotes refinement initialized with planar depth rather than NormDepth output.}
    \vspace{-10pt}
    \label{tab:cross_val}
\end{table}

\vspace{-10pt}
\paragraph{Dataset} 
To train the normal-based depth initialization network, as well as the self-supervised baseline Monodepth2~\cite{monodepth2}, we collected videos from $85$ clinical procedures and randomly sampled $185$k frames as the training set and another $5$k for validation.
We used the Colonoscopy 3D Video Dataset (C3DV)~\cite{bobrow2022} for the normal-depth refinement module, which provides ground truth depth maps and camera poses from a colonoscopy of a silicone colon phantom. We divide this dataset into 5 randomly-drawn cross-validation partitions with 20 training and 3 testing sequences such that the test sequences do not overlap.
The results reported in Section \ref{sec:ex-quant} are the methods' average performance across all folds.

\vspace{-10pt}
\subsection{Frame-wise Depth Evaluation} \label{sec:ex-quant}

\begin{figure}[t]
    \centering
    \includegraphics[width=.9\linewidth]{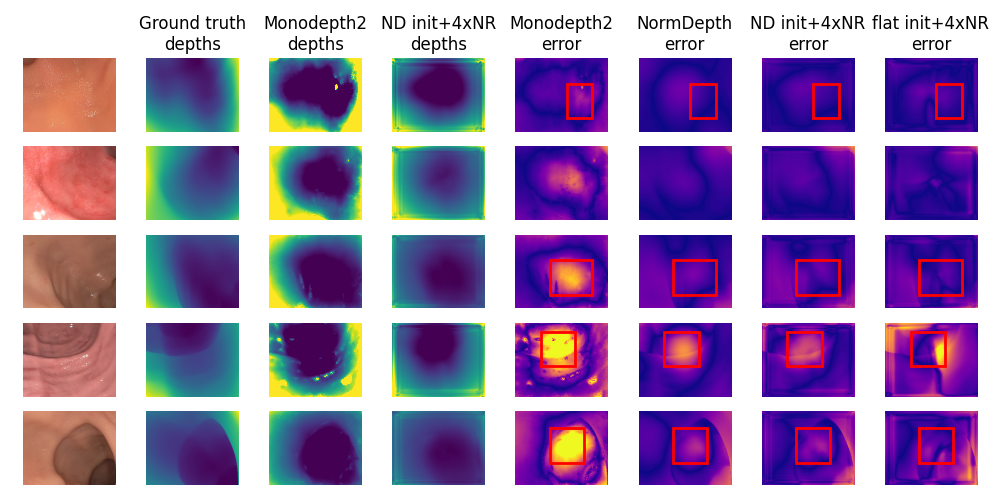}
    \vspace{-7pt}
    \caption{Example depth predictions and RMSE from C3DV. For the depth maps, darker colors denote more distant depths. For the RMSE, brighter colors denote higher error. Some areas of improvement are highlighted in boxes.}
    \label{fig:depth_errors}
    \vspace{-18pt}
\end{figure}

We compared our method's depth prediction with several ablations and the baseline against the ground truth in C3DV.
Following the practice in Godard et al.~\cite{monodepth2}, we rescaled our depth output to match the median of the ground truth and reported $4$ pixel-wise aggregated error metrics in Table \ref{tab:cross_val}.

Comparing depth prediction errors (Fig. \ref{fig:depth_errors}), both models using our two-stage method significantly outperform the photometric-based baseline Monodepth2, demonstrating the merit of emphasising geometric features (specifically surface normals) in colonoscopic depth estimation.
Meanwhile, although each individual stage of our two-stage method (NormDepth and flat init$+$NR) already produces relatively good performance, our combined system performs even better and generates the best quantitative result on this dataset (from ND init $+\:1 \times$NR). 
Notice that although based on the results from ablation models, the normal consistency loss $L_{norm}$ and multi-iteration of normal refinement quantitatively do not boost performance here due to the nature of C3DV dataset, they are critical for generating better 3D reconstruction shapes (Sections \ref{sec:ex-qual} and \ref{sec:ex-clinical}).

\subsection{C3DV Reconstruction} \label{sec:ex-qual}

\begin{figure}[t]
    \centering
    \includegraphics[width=.85\linewidth,trim={0 5cm 0 0},clip]{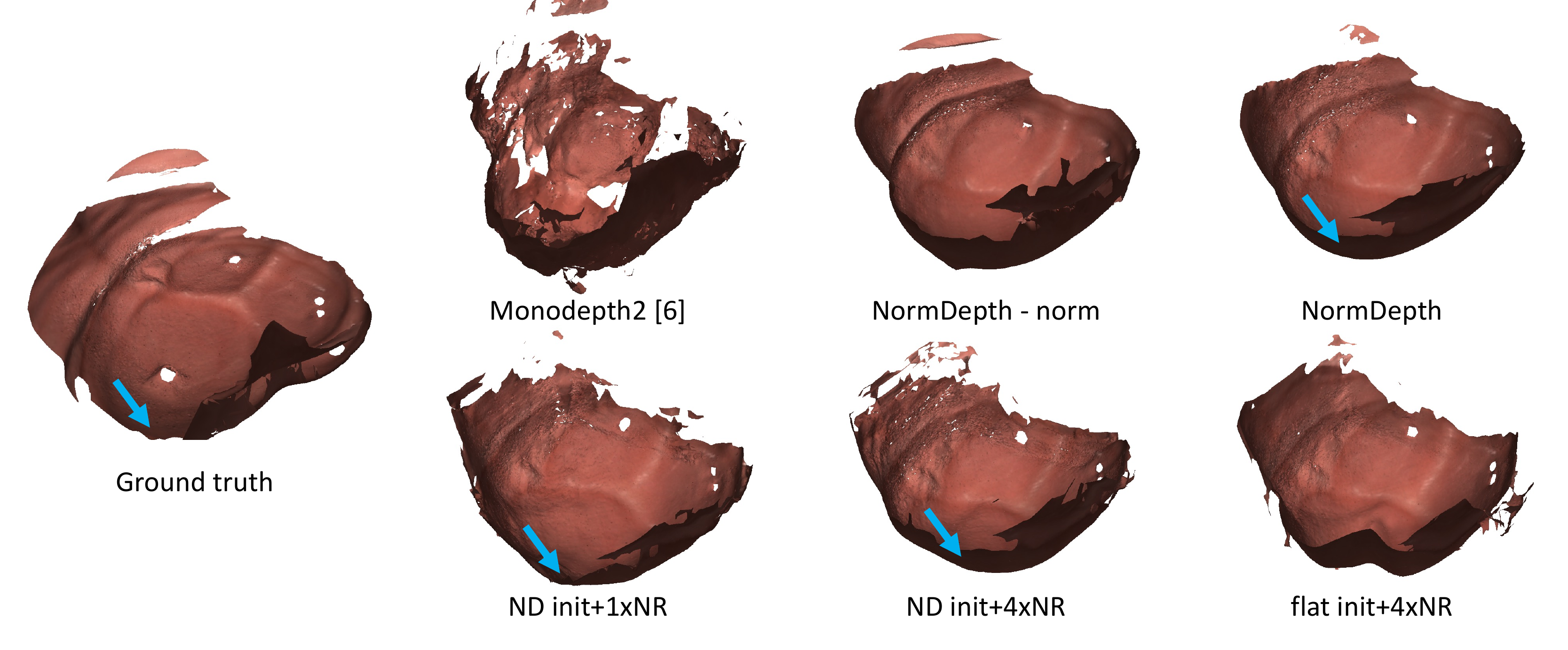}
    \includegraphics[width=.85\linewidth,trim={0 5.5cm 0 0},clip]{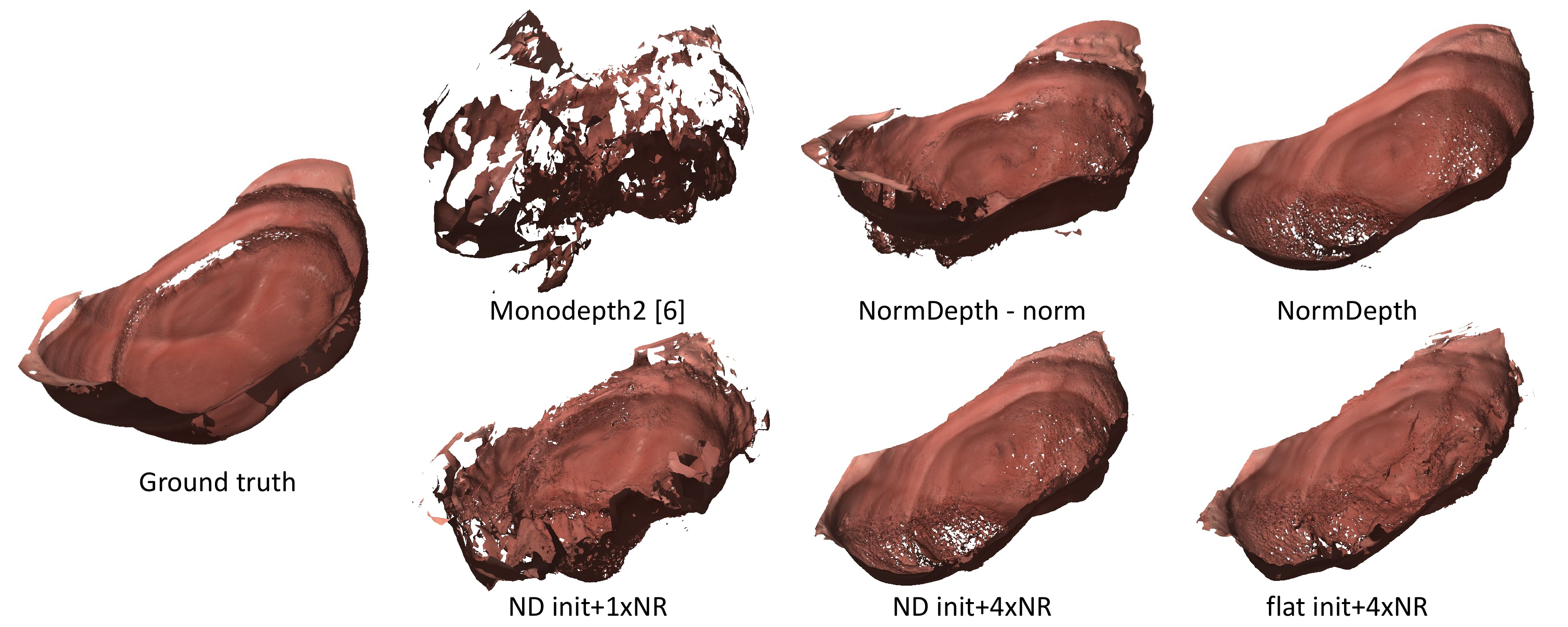}
    
    \vspace{-5pt}
    \caption{Example reconstructed sequences from C3DV using various methods of initialization for SLAM pipeline. The more planar shapes observed in the ND init$+n \times$NR compared to NormDepth variations are closer to the ground truth reconstruction while the noisy reconstructions using Monodepth2 and flat init$+4 \times$NR are farther from the ground truth. Select areas of improvement highlighted with arrows.}
    \label{fig:c4v3_recons}
    \vspace{-18pt}
\end{figure}

In this section, we demonstrate the improvement in reconstructions of the C3DV data using our normal-aware methods. In particular, we examine the effects of initializing our SLAM pipeline with the various depth and pose estimation methods.
Although C3DV provides a digital model of the phantom, here we compare against the reconstruction produced by using the ground truth depths and poses as initialization to our SLAM pipeline (and refer to this as the ground truth below). In this way, we can control for the impact of the SLAM pipeline in our reconstruction comparison. In Table \ref{tab:cross_val}, we measure the Chamfer distance from the ground truth to the reconstructed mesh after ordinary Procrustes alignment and optimizing the scaling factor for Chamfer distance from the ground truth to the reconstruction.

Overall, we find that the performance improvements observed in the frame-wise depth estimation are reflected in the reconstructions as well. Similarly, the weaknesses observed in the frame-wise inference also transfer to the reconstructions. In particular, we note that where significant noise is present in the frame-wise depth estimation for ND init+1xNR and reduced in ND init+4xNR, the corresponding reconstructions reflect the difference in noise as well.

In Fig. \ref{fig:c4v3_recons}, we visualize the reconstructions corresponding to example video sequences. In these sequences, we observe that our normal-aware methods significantly outperform the baseline Monodepth2 in qualitative similarity to the ground truth. In addition, we notice that the high curvature of the surface observed in the NormDepth and NormDepth-$L_{norm}$ reconstructions is reduced after refinement, bringing the overall reconstructed result closer to the ground truth.

\subsection{Clinical Reconstructions} \label{sec:ex-clinical}

\begin{figure}[ht!]
  \centering
  \resizebox{0.95\textwidth}{!}{
  \input{figs/real_sota/real_sota_fig_vert.tex}}
  \vspace{-5pt}
  \caption{3D reconstruction results on clinical colonoscopy data. Our combined system can handle both ``down-the-barrel'' and en face views, outperforming the photometric baseline Monodepth2. }
  \label{fig:real_sota}
  \vspace{-18pt}
\end{figure}
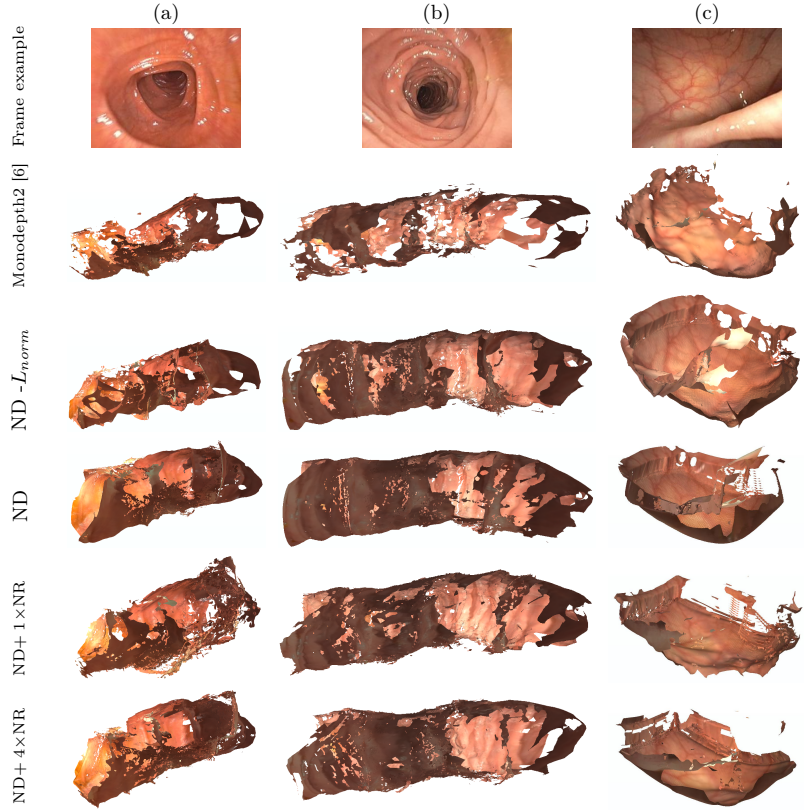

We tested our trained depth estimation models on clinical colonoscopy sequences and generated 3D reconstruction with the SLAM pipeline. 
Fig.~\ref{fig:real_sota} shows the reconstructed meshes of two ``down-the-barrel'' segments (Fig.~\ref{fig:real_sota}a and b) and an en face segment (Fig.~\ref{fig:real_sota}c).

The reconstruction quality from the two stages of our method (``ND'' and ``ND$+n \times$NR'') significantly outperforms the photometric baseline Monodepth2.
For ``down-the-barrel'' sequences where features are relatively rich, we expect a generalized cylinder shape with limited sparsity. For sequence (a), we expect two large blind spots due to occlusion by ridges and a slightly curved centerline. For sequence (b), we also expect two large blind spots due to the camera position but fairly dense surface coverage elsewhere. 
For these sequences, our predictions' shapes are more cylindrical and have surface coverage that more accurately reflect the quantity of surface surveyed compared to the reconstruction produced using Monodepth2.
The results also indicate that when trained without the normal consistency loss ($-L_{norm}$), NormDepth tends to predict more artifacts such as the skirt-shape outlier in sequence (a). This demonstrates the benefit of surface normal information in network training for improved consistency between frames.
Meanwhile, using multi-scale iterations of normal-depth refinement can reduce the noise and sparsity of reconstructed meshes compared to a single iteration.

For the en-face sequence (c), we expect a nearly planar surface.
Similar to the observations made in reconstructing sequences from C3DV, the high surface curvature produced from the initialization network is reduced after refinement, resulting in a more realistic reconstruction.

%% file: figs/real_sota/real_sota_fig_vert.tex
\newcommand{\FS}{0.15\columnwidth}
\newcommand{\md}{0.25\columnwidth}
\newcommand{\NDabla}{0.25\columnwidth}
\newcommand{\ND}{0.25\columnwidth}
\newcommand{\NRone}{0.25\columnwidth}
\newcommand{\NRfour}{0.25\columnwidth}
\newcommand{\threeone}{0.4\textwidth}
\newcommand{\onenine}{0.25\textwidth}
\newcommand{\enface}{0.25\textwidth}

\centering

\begin{tabular}{@{\hskip 0mm}c@{\hskip 6mm}c@{\hskip 2mm}c@{\hskip 2mm}c@{}}

& (a) & (b) & (c) \\

{\rotatebox{90}{\hspace{0mm}\scriptsize{Frame example}}} & 
\includegraphics[height=\FS]{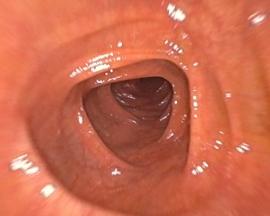} &
\includegraphics[height=\FS]{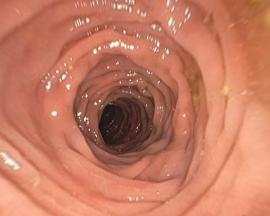} &
\includegraphics[height=\FS]{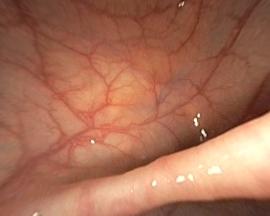} \\

{\rotatebox{90}{\hspace{-1mm}\scriptsize{Monodepth2~\cite{monodepth2}}}} & 
\includegraphics[width=\onenine]{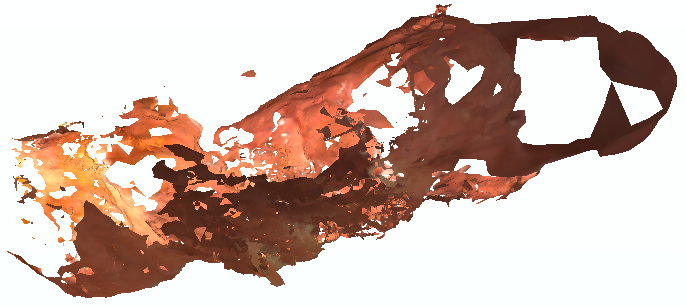} &
\includegraphics[width=\threeone]{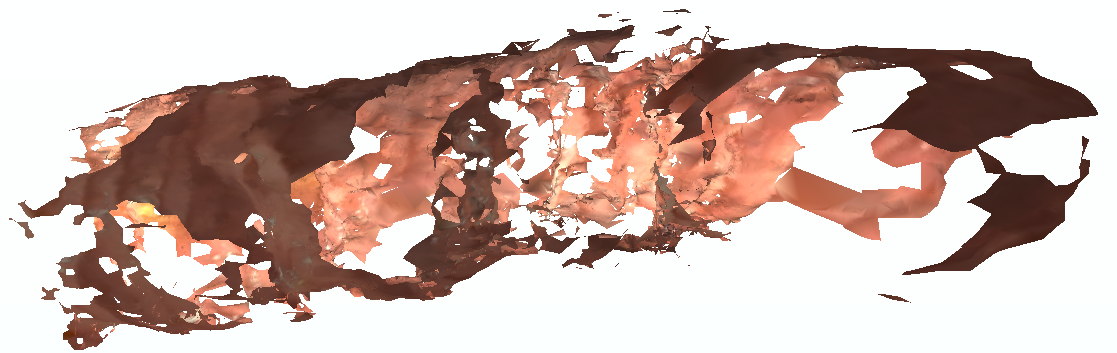} &
\includegraphics[width=\enface]{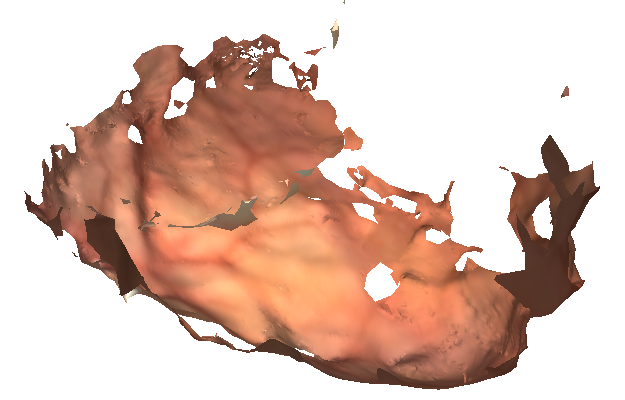} \\

{\rotatebox{90}{\hspace{0mm}\footnotesize{ND -$L_{norm}$}}} & 
\includegraphics[width=\onenine]{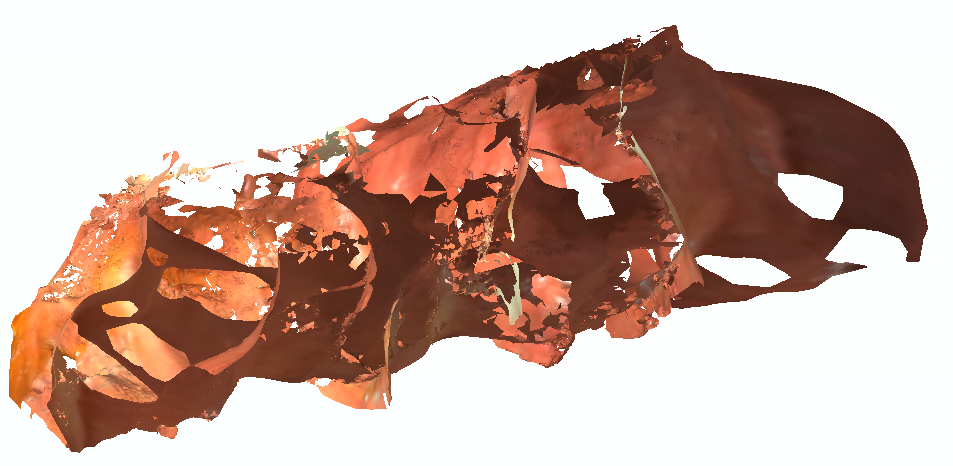} &
\includegraphics[width=\threeone]{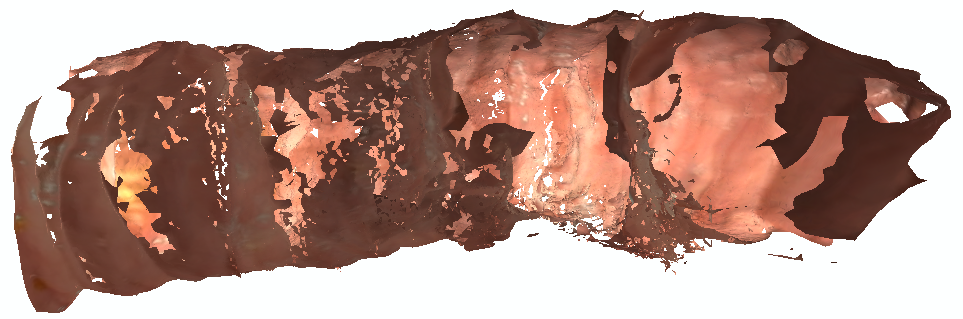} &
\includegraphics[width=\enface]{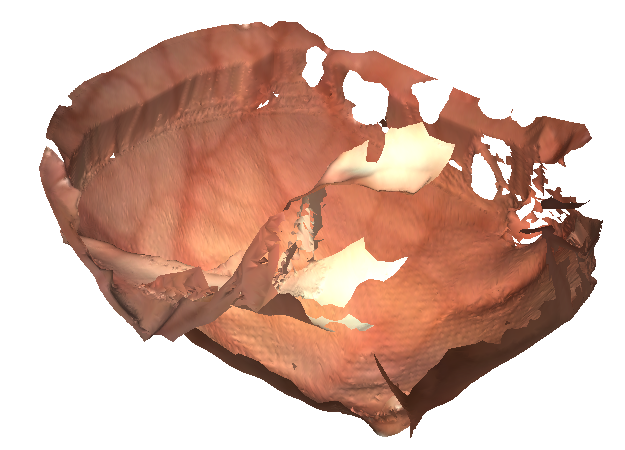} \\

{\rotatebox{90}{\hspace{4mm}\footnotesize{ND}}} & 
\includegraphics[width=\onenine]{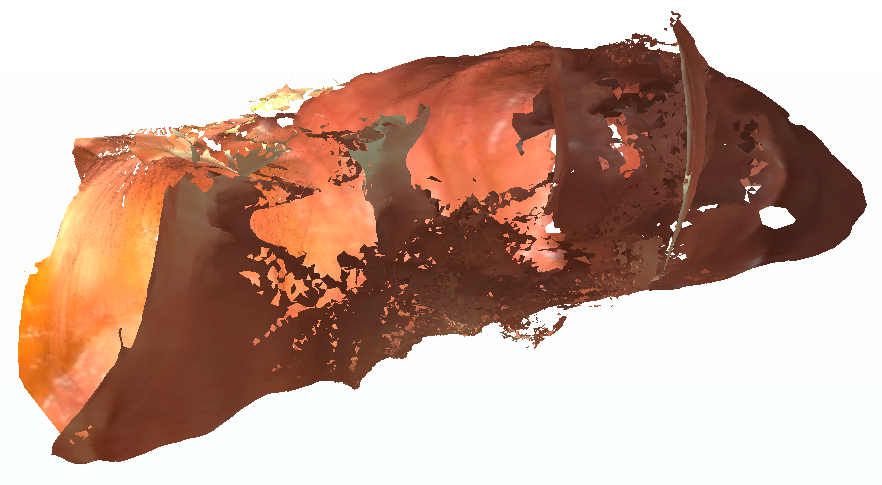} &
\includegraphics[width=\threeone]{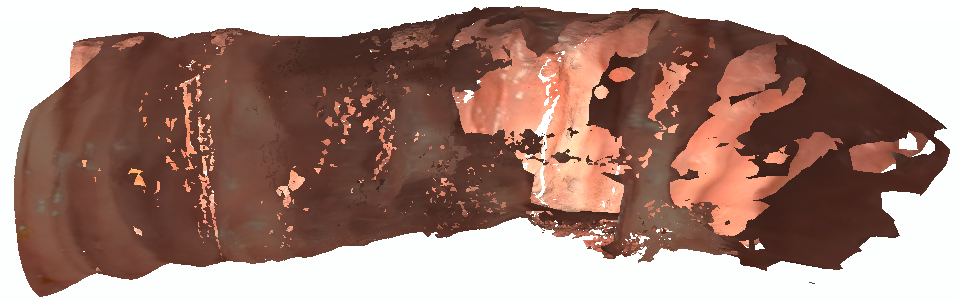} &
\includegraphics[width=\enface]{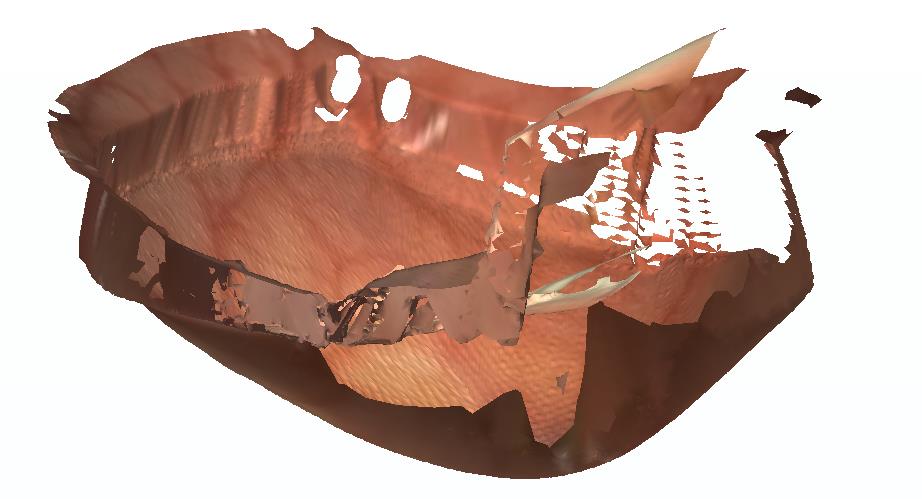} \\

{\rotatebox{90}{\hspace{1mm}\scriptsize{ND$+\:1 \times$NR}}} & 
\includegraphics[width=\onenine]{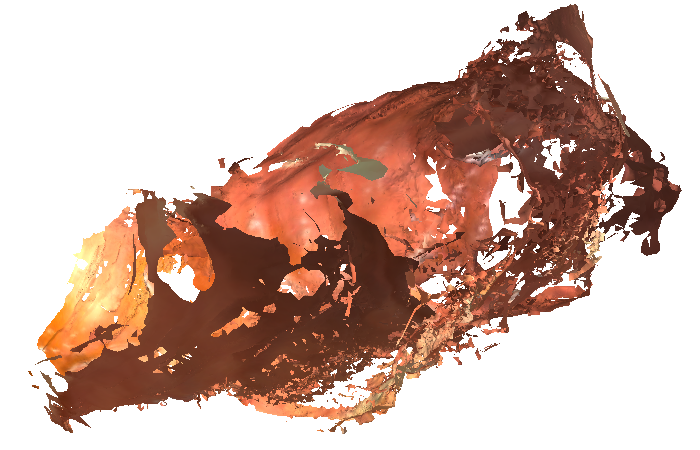} &
\includegraphics[width=\threeone]{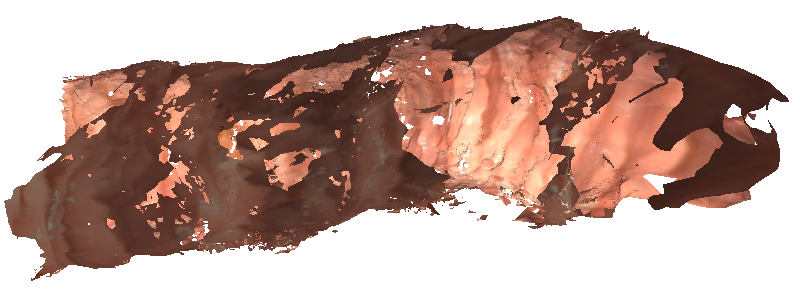} &
\includegraphics[width=\enface]{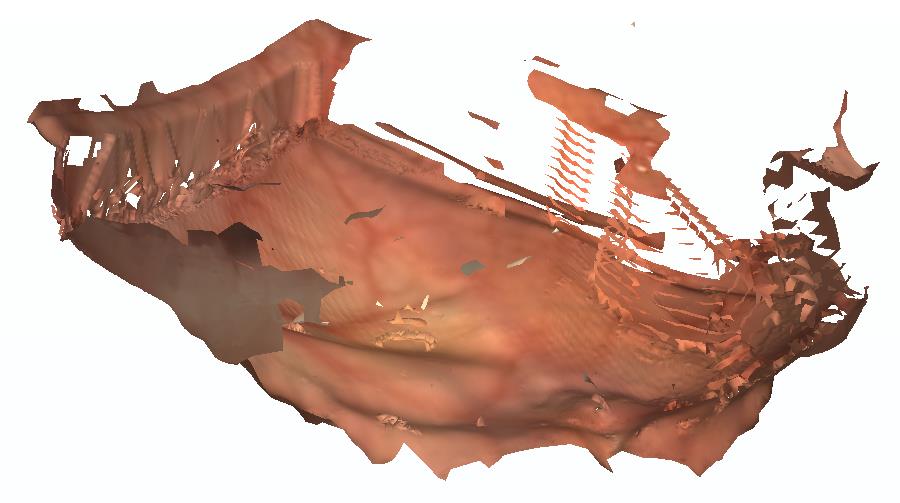} \\

{\rotatebox{90}{\hspace{1mm}\scriptsize{ND$+\:4 \times$NR}}} & 
\includegraphics[width=\onenine]{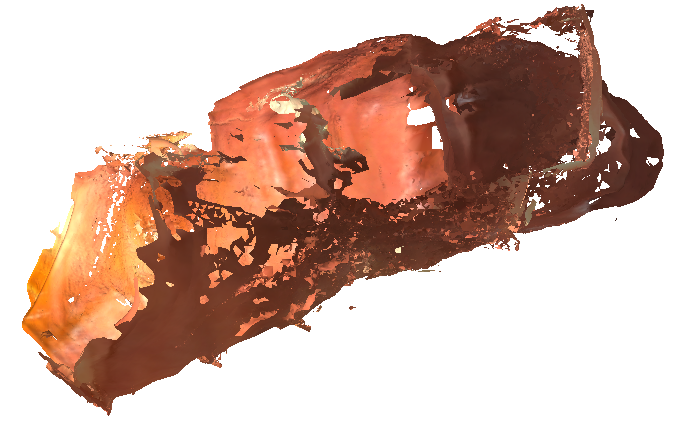} &
\includegraphics[width=\threeone]{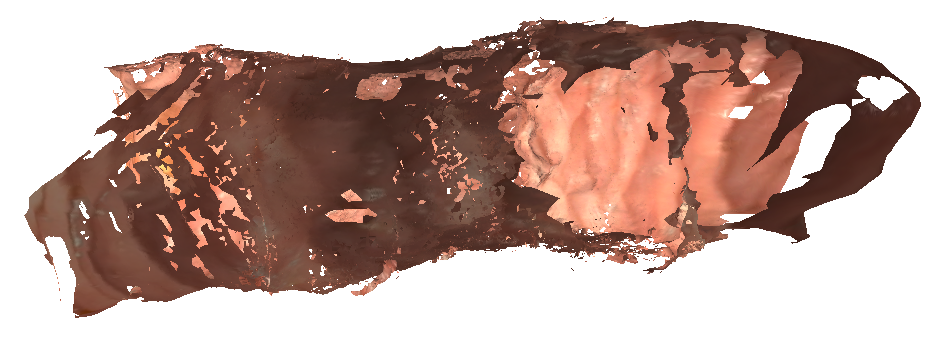} &
\includegraphics[width=\enface]{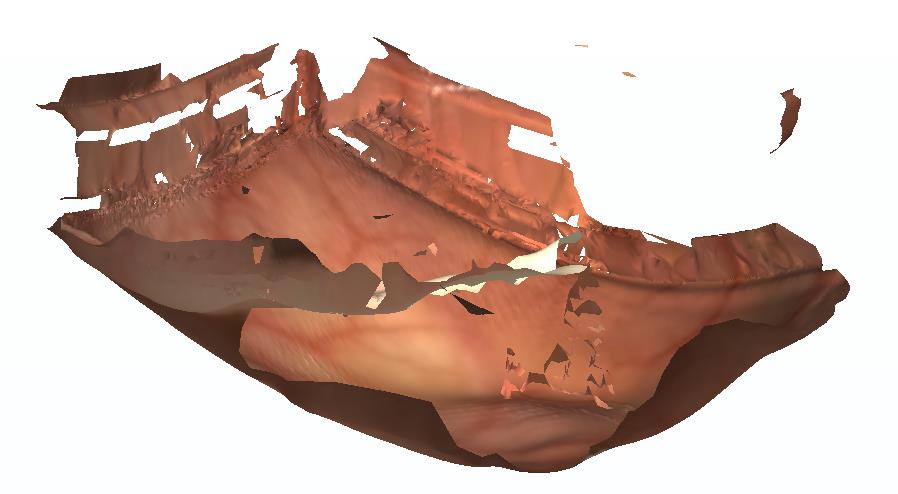} \\

\end{tabular}

%% file: 5_conclusion.tex
\section{Conclusion}
\vspace{-7pt}
In this work we introduced the use of surface normal information to improve frame-wise depth and camera pose estimation in colonoscopy video and found that this in turn improves our ability to reconstruct 3D surfaces from videos with low geometric texture. We used a combination of supervised and unsupervised losses to train our multi-stage framework and found significant performance improvements over methods that do not consider surface geometry. We have also shown that the incorporation of normal-aware losses allows us to reconstruct clinical videos of low-texture en face views.

\vspace{-8pt}
\paragraph{Limitations and Future Work}
In this work, we have treated "down-the-barrel" and en face views separately. In practice, colonoscopy videos transition between these two view types, so constructing a framework that can also transition between view types would have significant clinical application; we leave this investigation to future work.

\vspace{-8pt}
\paragraph{Acknowledgements}
We thank Zhen Li and his team at Olympus, Inc. for support and collaboration and Taylor Bobrow for early access to the C3DV dataset.

\vspace{-2pt}